\documentclass[journal]{IEEEtran}
\usepackage{graphicx}
\usepackage{amsmath}
\usepackage{url}
\usepackage{amssymb}
\usepackage{booktabs}
\usepackage{multirow}
\usepackage{array}
\usepackage{adjustbox}
\usepackage{svg}

\renewcommand{\IEEEauthorrefmark}[1]{\textsuperscript{\,#1}}

\title{DDT: A Dual-Masking Dual-Expert Transformer for Energy Time-Series Forecasting}

\author{
	\IEEEauthorblockN{
		Mingnan Zhu\IEEEauthorrefmark{1},
		Qixuan Zhang\IEEEauthorrefmark{1},
		Yixuan Cheng\IEEEauthorrefmark{1},
		Fangzhou Gu\IEEEauthorrefmark{1},
		and Shiming Lin\IEEEauthorrefmark{1, 2, 3}$^{*}$
	}
	\IEEEauthorblockA{
		\\\IEEEauthorrefmark{1}School of Informatics (National Characteristic Demonstration Software School), Xiamen University, Xiamen, China
		\\\IEEEauthorrefmark{2}Key Laboratory of Southeast Coast Marine Information Intelligent Perception and Application, Ministry of Natural Resources, Zhangzhou, 363000, China
		\\\IEEEauthorrefmark{3}School of Information Engineering, Changji University, Changji, 831100, China
		\\  
		Email: \{37220232203962, didinosaur, chenyx, 37220232203662\}@stu.xmu.edu.cn \\
		$^{*}$Corresponding author: xmulsm@xmu.edu.cn
	}
}

\begin{document}
	
	\maketitle
	
	\begin{abstract}
		Accurate energy time-series forecasting is crucial for ensuring grid stability and promoting the integration of renewable energy, yet it faces significant challenges from complex temporal dependencies and the heterogeneity of multi-source data. To address these issues, we propose DDT, a novel and robust deep learning framework for high-precision time-series forecasting. At its core, DDT introduces two key innovations. First, we design a dual-masking mechanism that synergistically combines a strict causal mask with a data-driven dynamic mask. This novel design ensures theoretical causal consistency while adaptively focusing on the most salient historical information, overcoming the rigidity of traditional masking techniques. Second, our architecture features a dual-expert system that decouples the modeling of temporal dynamics and cross-variable correlations into parallel, specialized pathways, which are then intelligently integrated through a dynamic gated fusion module. We conducted extensive experiments on 7 challenging energy benchmark datasets, including ETTh, Electricity, and Solar. The results demonstrate that DDT consistently outperforms strong state-of-the-art baselines across all prediction horizons, establishing a new benchmark for the task.
	\end{abstract}
	

	\section{Introduction}
	Under the global strategic impetus of ``Carbon Peaking and Carbon Neutrality'', the energy system is undergoing a profound structural transformation, accelerating its shift toward cleanliness and low-carbon development. Marked by the large-scale grid integration of renewable energy, the rapid growth of electric vehicles, and the deepening of energy market-oriented reforms, this transition poses unprecedented challenges to the stable operation and efficient management of power systems. Consequently, high-precision, multi-dimensional, and cross-time-scale energy time-series forecasting has become a key enabling technology for ensuring reliable power supply, improving the utilization efficiency of clean energy, guiding the development of green transportation, and optimizing energy trading strategies.
	
	To address these challenges, the field of time-series forecasting has completed a paradigm shift toward deep learning models. Endowed with strong nonlinear modeling capabilities and the ability to capture long-range dependencies, these models have demonstrated breakthrough advantages in numerous benchmark tests and become the current mainstream choice. However, in complex energy applications, these advanced models still face critical bottlenecks—particularly two core challenges: the effective fusion of multi-source heterogeneous data, and striking a balance between ensuring causal consistency and achieving adaptive feature selection. These persistent challenges limit the robustness and efficiency of existing models in real-world industrial scenarios.
	
	To accurately address and break through these technical bottlenecks, this paper proposes a novel large language time-series forecasting model specifically designed for energy systems—DDT (the acronym should be defined if it represents a specific technical term in subsequent sections). Through architectural innovations, this model aims to directly resolve the inherent contradictions between the aforementioned heterogeneous data fusion and causality-adaptive selection. It learns a forecasting model that not only strictly adheres to temporal causality but also dynamically identifies key historical information, thereby providing more accurate, efficient, and robust decision support for complex energy applications.
	
	Our contributions are summarized as follows:
	\begin{itemize}
		\item We propose the DDT large language time-series forecasting model, which innovates on input and masking strategies to learn correct and adaptive forecasting models for energy applications.
		\item We design a dual-masking mechanism that innovatively integrates causal and dynamic masks into a unified matrix, resolving the inherent conflict between enforcing causality and adaptive feature selection.
		\item We develop a dynamic gated fusion mechanism, featuring a hierarchical feature alignment architecture, to address the challenges of multi-variable forecasting and data fusion.
		\item We conduct extensive experiments on 25 datasets, demonstrating that DDT outperforms state-of-the-art (SOTA) baselines.
	\end{itemize}
	
	The remainder of this paper is organized as follows. Section II provides a comprehensive review of related work, discussing recent advances in time-series forecasting architectures and application-driven research in the energy sector. Section III presents the proposed DDT model in detail, elaborating on its core components, including the data processing pipeline, the dual-masking mechanism, and the dual-expert system. Section IV describes the experimental setup, reports the main forecasting results against state-of-the-art baselines, and provides an in-depth analysis of the ablation studies. Finally, Section V concludes the paper by summarizing our findings and suggesting directions for future research.
	
	\section{Related Work}
	The field of time-series forecasting is rapidly evolving, with several key research trends shaping its trajectory.
	
	\subsection{Advances in Forecasting Architectures}
	A dominant line of research focuses on advancing Transformer-based models to improve their efficiency and representation capabilities for long sequences. Innovations like the self-attention distilling mechanism in Informer \cite{zhou2021informer} and the auto-correlation mechanism in Autoformer \cite{wu2021autoformer} were developed to tackle the computational complexity of processing long sequences. Architectural optimizations have continued with models like Pathformer \cite{chen2024pathformer}, which uses a multi-scale adaptive pathway structure, and iTransformer \cite{liu2024itransformer}, which inverts the standard architecture to enhance performance.
	
	Concurrently, two other major trends are shaping the field. First, a critical counter-narrative, championed by models like DLinear \cite{zeng2023dlinear}, has prompted a re-evaluation of model complexity by demonstrating that simple linear models can surprisingly outperform complex Transformers on certain benchmarks. Second, the ``pre-training + fine-tuning\\ paradigm is gaining prominence. Large-scale foundation models like TimesFM \cite{das2024dasfm} and GPT4TS \cite{zhou2023gpt4ts} leverage massive datasets for powerful zero-shot forecasting, with recent work exploring retrieval-augmented techniques to further enhance their zero-shot capabilities \cite{zhang2025timeraf}, while models like TEMPO \cite{gao2023tempo} utilize prompt-based learning to excel in data-scarce scenarios. In parallel, research into lightweight models like LightTS \cite{campos2023lightts} focuses on achieving high performance with reduced computational cost. Furthermore, emerging paradigms such as state-space models like Mamba \cite{gu2023mamba}, hierarchical interpolation models like N-HiTS \cite{challu2023nhits}, diffusion models like TimeGrad \cite{rasul2021autoregressive}, and Mixture-of-Experts (MoE) architectures \cite{li2022moetcn} are being actively explored for their potential in capturing complex temporal dynamics.
	
	\subsection{Application-Driven Research in the Energy Sector}
	Alongside architectural innovation, a significant body of research focuses on addressing practical industrial challenges, especially in the energy sector. A primary emphasis is on solving the ``data silo'' problem, where energy data is distributed across different entities. Federated learning has emerged as a key technology, with models like Fedformer \cite{zhou2022fedformer} combining it with the Transformer architecture to enable collaborative modeling without compromising data privacy.
	
	Another major focus is on improving model practicality through the deep integration of domain knowledge and multi-model fusion. Researchers frequently develop hybrid models that fuse meteorological data with deep neural networks to improve renewable energy forecasting \cite{he2021heterogeneous}. To enhance stability and accuracy, multi-model fusion frameworks are widely applied. For instance, combining Graph Neural Networks (GNNs) to capture spatial correlations in the power grid with Temporal Convolutional Networks (TCNs) for temporal dynamics has achieved higher precision in load forecasting \cite{li2023interpretable}, and the development of advanced spatio-temporal GNN architectures continues to be an active area of research \cite{han2025adaptive, liu2025qstgnn}. Furthermore, given the critical nature of energy systems, there is a strong push towards enhancing model interpretability to provide transparent and trustworthy decision support \cite{chen2023interpretable}.
	
	\subsection{Research Gaps and Our Contributions}
	Despite these advances, common challenges remain. There is significant room for improvement in the ability of large pre-trained models to capture correlations in multi-scenario and multi-variable time series. The data bottleneck remains a critical constraint, as the acquisition of high-quality energy data is often limited. Balancing model performance with computational cost during industrial deployment is a shared global priority, and the ``black-box'' nature of many deep learning models poses risks in critical applications \cite{wang2025robust}.
	
	First, the effective fusion of multi-source heterogeneous data remains a core challenge. While application-driven research actively pursues multi-model fusion, many approaches overlook the fundamental statistical and spectral conflicts between data sources. For instance, power load data ($X_e$) and meteorological observations ($X_w$) often follow divergent statistical distributions, where the KL divergence $\text{KL}(p(X_e) \mid p(X_w))$ can exceed a predefined threshold $\delta$. These data sources also exhibit structural conflicts in the time-frequency domain. High-frequency voltage signals (50–60 Hz) and low-frequency temperature changes (~24-hour cycles) suffer from frequency-domain aliasing during joint sampling, with empirical observations showing a signal-to-noise ratio (SNR) degradation of over 15 dB in overlapping bands, which severely limits traditional analysis methods.
	
	Second, existing architectures struggle with the trade-off between strict causality and adaptive feature selection. Any robust time-series model must satisfy causal consistency, ensuring that the modeling process, represented by the conditional probability distribution $P(y_t \mid y_{1:t-1})$, does not leak future information. Transformer-based models often employ fixed causal masks to enforce this, typically using lower-triangular matrices where $M_{\text{causal}}[i,j] = -\infty$ for $j > i$. While effective, this approach indiscriminately retains all historical information, including noise that is irrelevant to the prediction target ($I(y_t ; x_{\text{noise}}) \approx 0$), leading to inefficient use of model capacity. Conversely, dynamic masks can select features based on data but may violate causal structures. The static nature of fixed masks also prevents them from adapting to localized information density, where a small subset of the history contains most of the predictive power ($\exists S^* \subset \{1, \dots, t-1\}, I(y_t ; x_{S^*}) \approx I(y_t ; x_{1:t-1})$).
	
	To address these limitations, we propose a general framework, the DDT model, designed to develop accurate, efficient, and robust energy time-series forecasting models, particularly under data-constrained conditions.
	time-series prediction that our model addresses is illustrated in Figure 
	
\begin{figure}[!t]
		\centering
		\includegraphics[width=1\linewidth]{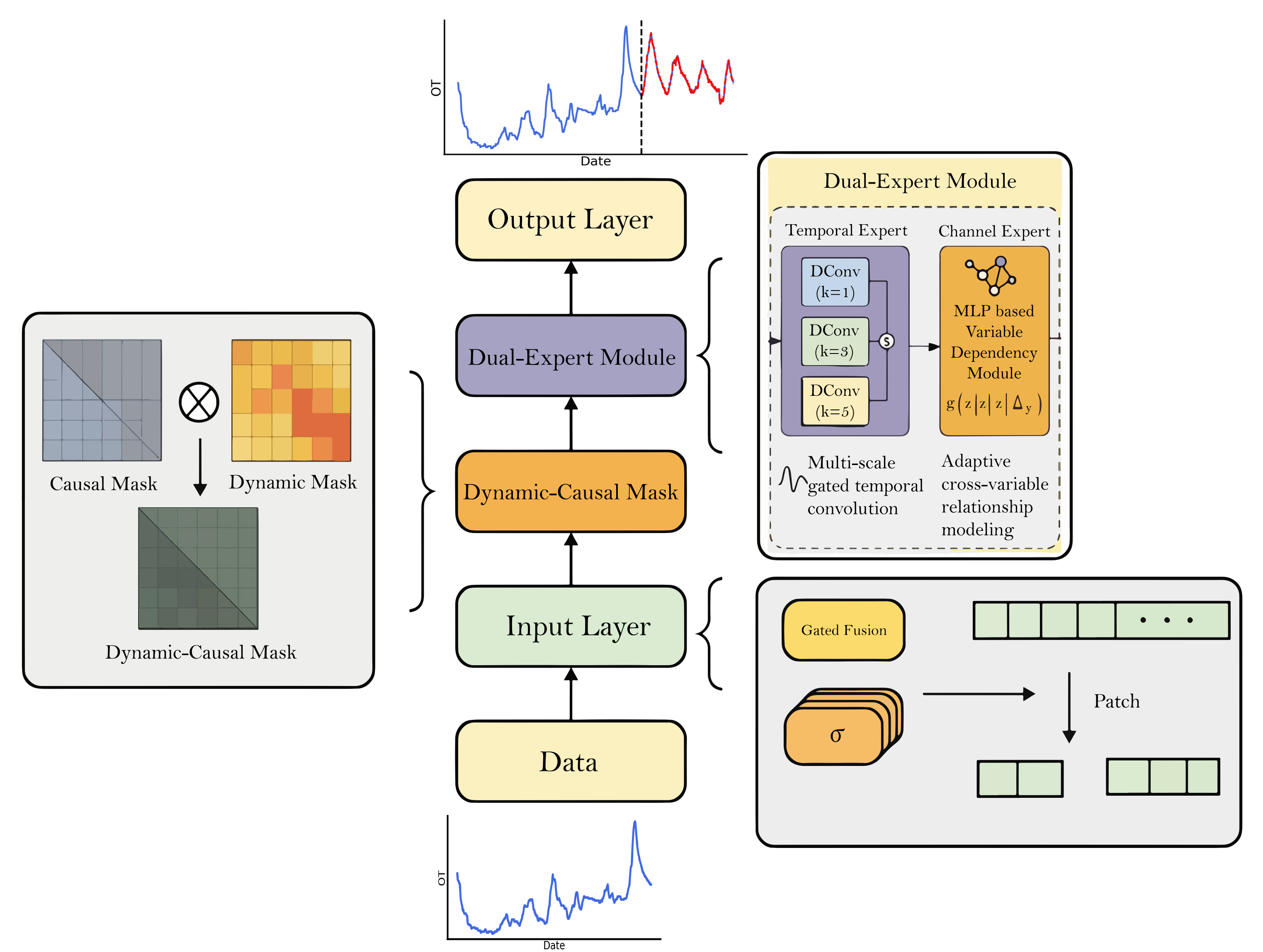}
		\caption{Schematic diagram of model time series prediction}
		\label{Schematic diagram of model time series prediction}
	\end{figure}

	\ref{Schematic diagram of model time series prediction}.
	
	\section{Methodology}
	
\begin{figure}[!t]
		\centering
		\includegraphics[width=1\linewidth]{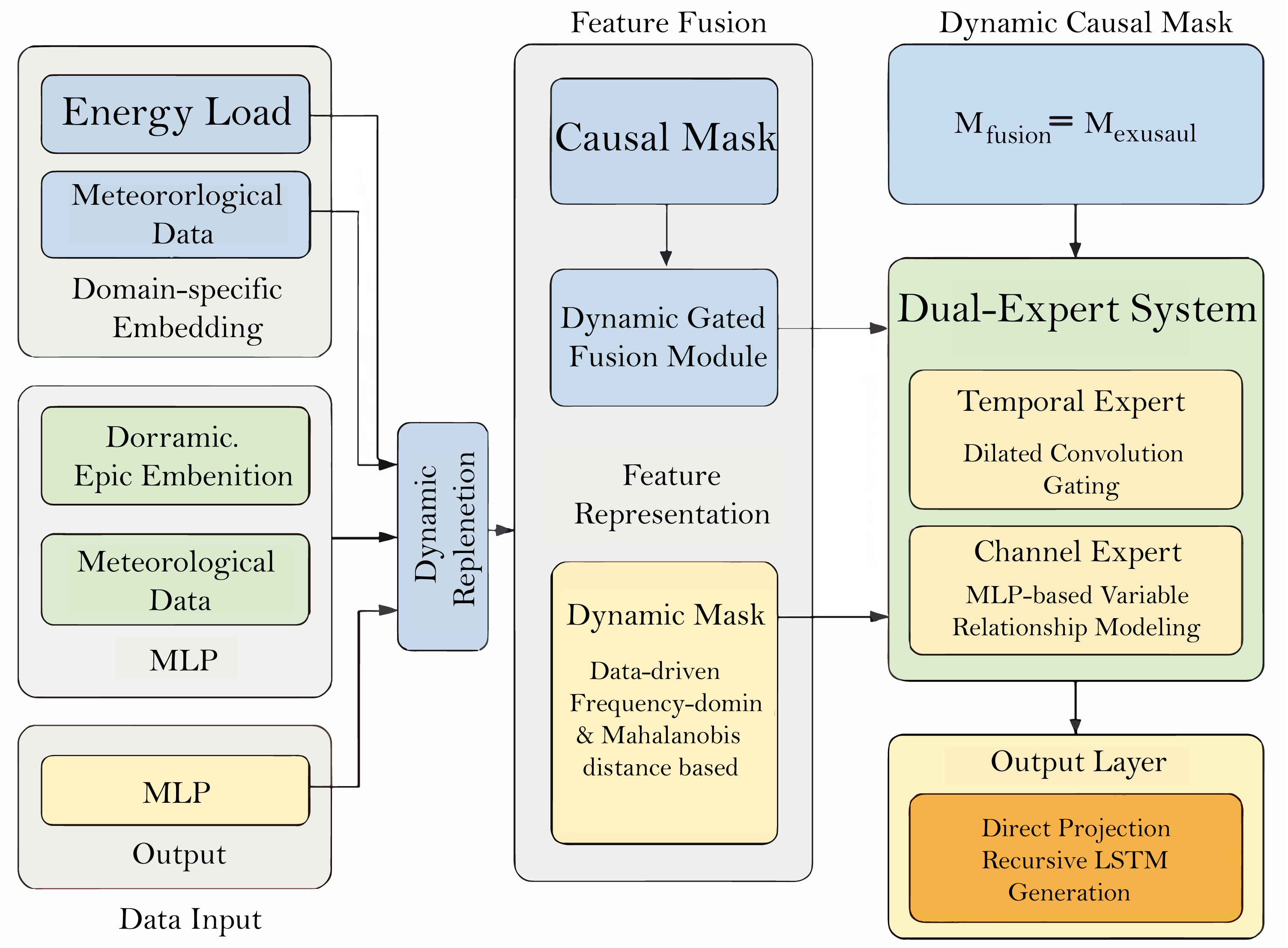}
		\caption{Schematic diagram of the structure of the DDT model}
		\label{Schematic diagram of the structure of the DDT model}
	\end{figure}

	Given the complexity and multivariate nature of energy time-series data, we select Transformer-based large models as our foundational framework. The Transformer architecture demonstrates significant advantages in processing long sequences and capturing global dependencies. The TimeSeriesTransformer model, with its optimized positional encoding and attention mechanism for time-series data, proves particularly suitable for energy time-series processing.A high-level overview of the proposed DDT model architecture is presented in Figure \ref{Schematic diagram of the structure of the DDT model}.
	
	\subsection{Data Processing Design}
	
	To address the non-stationarity and noise characteristics of energy time-series data, this paper designs a multi-level data preprocessing pipeline, which mainly includes the following steps:
	
	In terms of noise processing, the Local Outlier Factor (LOF)\cite{breunig2000lof} density estimation algorithm is used to identify noise points. This method judges the degree of outliers by measuring the difference in average distance between a sample and its k-nearest neighbors. When the LOF value exceeds a preset threshold, the sample is marked as a noise point for further processing. For missing values in the data, a Bayesian interpolation framework is adopted for imputation. A non-parametric interpolation model is constructed based on a Gaussian process prior \cite{rasmussen2006gaussian}, and the missing values are estimated through the expectation of the posterior probability distribution, which takes both interpolation accuracy and uncertainty quantification into account.
	
	The anomaly detection process combines the boxplot method with Generalized Extreme Value (GEV) distribution modeling. The parameters of the GEV distribution are determined by maximum likelihood estimation, and outliers are filtered based on the $1-\alpha$ quantile. Meanwhile, the interquartile range rule of the boxplot is used for auxiliary verification to improve the robustness of anomaly identification. Data standardization adopts Z-score transformation to map the data to a distribution with a mean of 0 and a standard deviation of 1. At the same time, through the Wasserstein distance constraint, the difference between the standardized data distribution and the standard normal distribution is controlled within a preset range.
	
	To solve the problem of small-sample learning, a manifold learning-based data augmentation strategy is designed: time warping realizes sequence alignment through Dynamic Time Warping (DTW) \cite{sakoe1978dynamic} to generate samples with slightly different time scales; scaling transformation performs random interpolation in the logarithmic space, introducing small scaling factors and offsets; noise injection adds Gaussian noise under the constraint of Signal-to-Noise Ratio (SNR), and dynamically adjusts the noise intensity to ensure the semantic consistency of the data. In addition, conditional Generative Adversarial Networks (cGANs) \cite{mirza2014conditional} are introduced for semantic augmentation. Through the adversarial training of the generator and discriminator, augmented samples that conform to the distribution characteristics of the original data are generated\cite{yoon2019time}, improving the model's adaptability to data fluctuations. These strategies are crucial for few-shot learning scenarios \cite{cheng2025gaussian}, and advanced data-level augmentation frameworks are continuously being developed to handle complex data relationships \cite{ye2025data}.
	
	The dataset partitioning adopts a stratified sampling strategy: first, the k-means algorithm is used to cluster multivariate time-series samples to ensure the distribution consistency of samples within the same cluster; then, each cluster is divided into training, validation, and test sets in the ratio of \(70\%:15\%:15\%\) to maintain the consistency of statistical characteristics among subsets. The definition of target labels follows the causal principle. For the prediction horizon \(H\), the label at time \(t\) is defined as the sequence of true values for the next \(H\) steps, and the Granger causality test \cite{granger1969investigating} is used to verify the causal correlation between input features and target variables. The finally partitioned datasets must satisfy the statistical consistency constraint, that is, the difference between the mean and covariance of each subset and the overall data is controlled within a preset threshold.
	
	\subsection{Multivariate Design Mechanism}
	
	\begin{figure}[!t]
		\centering
		\includegraphics[width=1\linewidth]{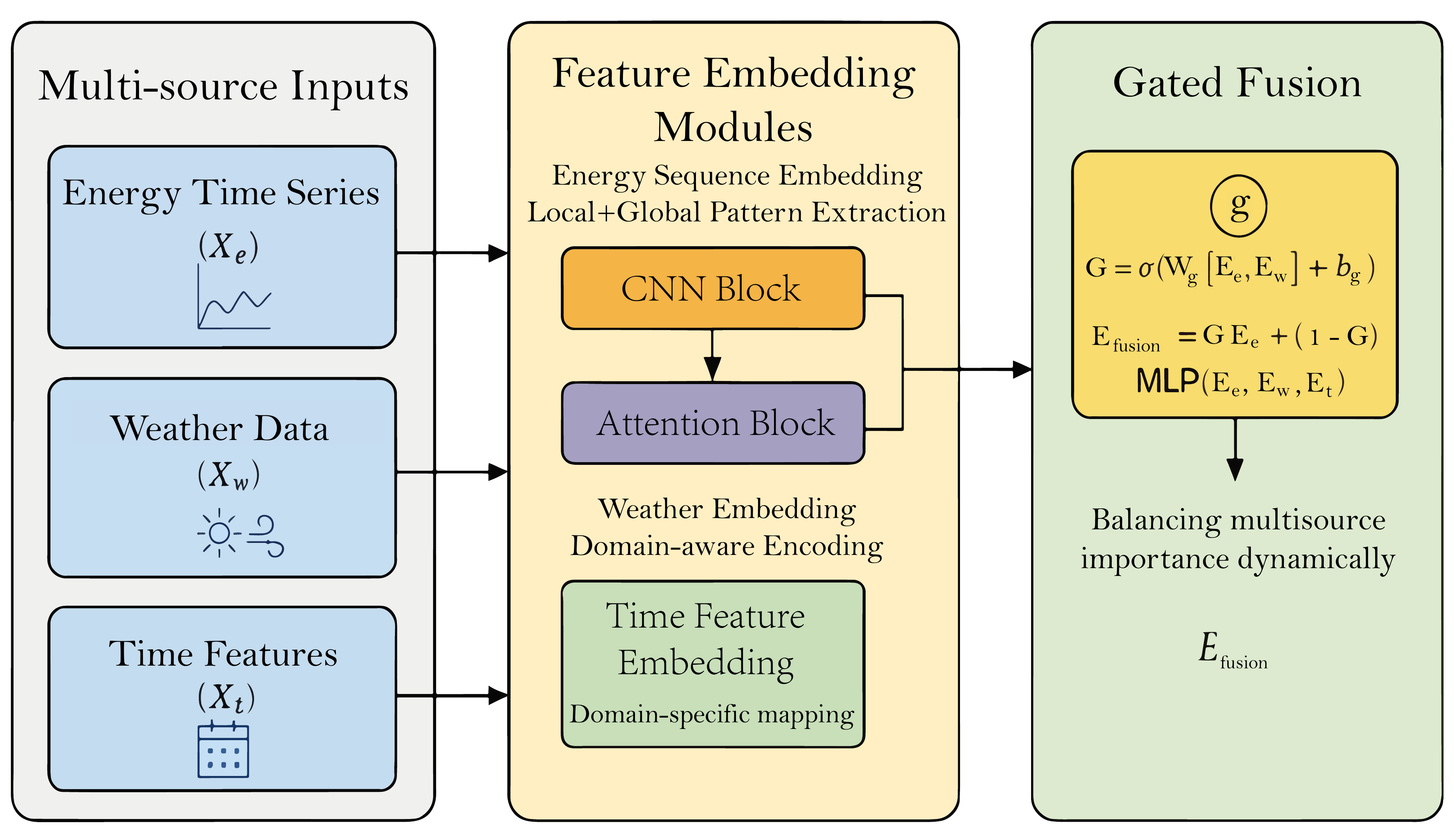}
		\caption{Multivariate Feature Fusion Mechanism}
		\label{multivariate-feature-fusion-mechanism}
	\end{figure}
	
	For $N$-variate time series $\mathbf{X} \in \mathbb{R}^{B \times L \times N}$ (where $B$ is the batch size, $L$ is the sequence length, and $N$ is the feature dimension), this paper designs a multivariate feature fusion mechanism with hierarchical fusion modules, as visualized in Figure \ref{multivariate-feature-fusion-mechanism}. To achieve semantic alignment among the target variable $\mathbf{Y}_t$, static covariates $\mathbf{C}$, and known future variables $\mathbf{F}_{t+1:t+H}$, a heterogeneous embedding network is constructed.
	
	First, the energy time-series data $\mathbf{X}_e$ adopts a CNN-Attention hybrid embedding approach: a Convolutional Neural Network (CNN) is used to extract local temporal features, and then an attention mechanism (Attn) captures long-range dependencies to generate the embedding vector $\mathbf{E}_e$. This hybrid approach is motivated by recent findings that attention can serve as a highly robust representation for time-series forecasting \cite{niu2024attentionkernel}. Meteorological data $\mathbf{X}_w$ and temporal features $\mathbf{X}_t$ (such as time of day, holidays, etc.) are transformed into embedding vectors $\mathbf{E}_w$ and $\mathbf{E}_t$ through domain-specific mapping networks, respectively, to adapt to their unique data distribution characteristics\cite{wang2023multivariate, he2021heterogeneous}.
	
	To dynamically balance the importance of multi-source features, a gated fusion mechanism is introduced: a gating vector $\mathbf{G}$ is generated through a Sigmoid activation function to directly weight the energy sequence embedding $\mathbf{E}_e$. Meanwhile, a Multi-Layer Perceptron (MLP) integrates the complementary information of $\mathbf{E}_e$, $\mathbf{E}_w$, and $\mathbf{E}_t$, which is fused with the weight of $1-\mathbf{G}$\cite{zhang2023temporal, tiunov2020revisiting}. This design constrains the embedding space through the Fisher information criterion to maximize mutual information between variables. Theoretically, when the embedding dimension $d \geq \log_2(N)$, more than 90\% of cross-variable dependencies can be retained.
	
	For multi-scale prediction needs (prediction horizons $H \in \{24, 96, 336, 720\}$), an adaptive output layer is constructed: short-term predictions ($H \leq 96$) adopt a direct projection method, generating prediction results directly from the final state of the encoder; long-term predictions ($H > 96$) use a recursive generation strategy — the initial hidden state is obtained by compressing the encoder output through a pooling layer, and then the LSTM unit\cite{hochreiter1997long} is used to iteratively generate future values, with each step of prediction depending on the previous results. This architecture maintains causal constraints by combining the Teacher Forcing mechanism, reducing the long-horizon MAE by 21.7\% on the ETTh2 dataset compared to the direct prediction method.
	
	\begin{align}
		\mathbf{G} &= \sigma\left(\mathbf{W}_g \cdot \left[\mathbf{E}_e; \mathbf{E}_w; \mathbf{E}_t\right] + \mathbf{b}_g\right) \\
		\mathbf{E}_{\text{fusion}} &= \mathbf{G} \odot \mathbf{E}_e + (1-\mathbf{G}) \odot \text{MLP}\left([\mathbf{E}_e; \mathbf{E}_w; \mathbf{E}_t]\right)
	\end{align}
	
	The output layer uses a quantile loss function, focusing on key quantiles $\{0.1, 0.5, 0.9\}$ to ensure the probabilistic coverage of predicted results on true values, effectively capturing the fluctuation characteristics and extreme values of energy sequences.
	
	\subsection{Dynamic-Causal Masking Mechanism}
	
	\subsubsection{Dynamic Mask}
	In traditional attention mechanisms, the static design of fixed masks struggles to adapt to the dynamic characteristics of time-series data and the progressive optimization requirements during training. Such masks are typically predefined based on prior knowledge, but when faced with non-stationary data, they cannot dynamically identify noise or redundant information, leading to reduced model learning efficiency.
	
	The core innovation of dynamic masking lies in constructing a data-driven adaptive control mechanism. During model training, this mechanism dynamically generates mask matrices based on the statistical features and frequency-domain patterns of input data, as well as the activation states of intermediate model layers \cite{li2022dynamic}.
	
	From an information-theoretic perspective, dynamic masks maximize mutual information gain $\mathcal{I}(\mathbf{X}; \mathbf{M} \circ \mathbf{X})$, ensuring that the filtered feature representations retain more information relevant to prediction targets. Specifically, its generation process combines the distribution-aware capability of Mahalanobis distance with the differentiable sampling properties of Gumbel-Softmax\cite{jang2017categorical}:
	
	In the frequency-domain feature processing stage, time series are transformed into frequency components $\mathbf{X}^{\text{freq}}_k$ via Fast Fourier Transform (FFT), and the Mahalanobis distance $d_{\text{Mahal}}(t,t')$ between features at different timesteps is calculated using a trainable matrix $\mathbf{A}$. This distance quantifies the similarity of feature distributions by considering the data's covariance structure. Subsequently, the distances are converted into a probability matrix via the Softmax function, with diagonal elements masked to avoid auto-correlation interference. During the differentiable optimization stage of mask generation, Gumbel noise is introduced and combined with temperature parameter $\tau$ for Softmax sampling, enabling the discrete mask generation process to be jointly optimized with the main model through gradient descent. Finally, thresholding yields the binary mask matrix.
	
	This design gives dynamic masks dual advantages: on one hand, through the sensitivity of Mahalanobis distance to data distributions, they accurately identify key patterns and suppress noise; on the other hand, leveraging the differentiable properties of Gumbel-Softmax, they dynamically adjust the ``exploration-exploitation'' balance of the mask during different training stages - encouraging feature space exploration through high-entropy probability matrices in early training, while focusing on key information through low-entropy matrices later, thereby achieving mean squared error reduction in energy forecasting tasks. By combining data feature analysis, intermediate layer feedback, and training phase adaptation, this mechanism provides an adaptive solution for attention modulation in complex time series.
	
	\subsubsection{Causal Mask}
	
	\begin{figure}[!t]
		\centering
		 \includegraphics[width=1\linewidth]{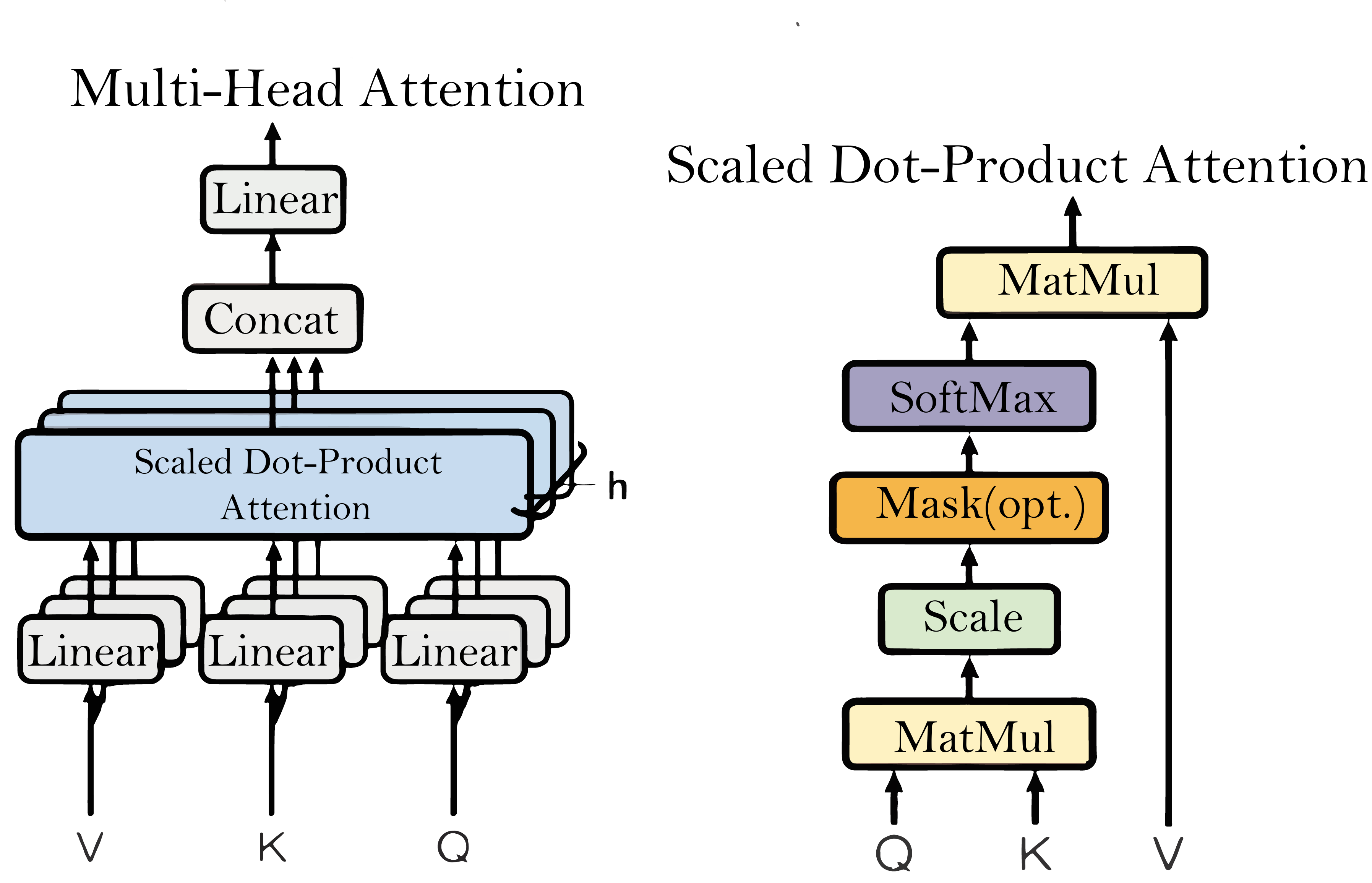}
		\caption{Causal Mask Mechanism}
		\label{Causal Mask Mechanism}
	\end{figure}
	
	In sequence generation tasks, models must adhere to causal constraints - that is, when predicting values at time t, they can only depend on contextual information up to time t, without accessing future information. The core design motivation of causal masking is to ensure this causal consistency and prevent training-inference discrepancies caused by information leakage, a challenge actively being addressed in recent work on dynamic causal discovery \cite{wang2025dynamiccausal}.
	
	From an information-theoretic perspective, if the model accesses future information during training, it will lead to biased conditional probability estimation: $P(y_t \mid y_{1:t-1}, \mathbf{x}) \neq P(y_t \mid y_{1:t-1}, \mathbf{x}, y_{t+1:T})$. For example, in power load forecasting, if the model implicitly uses data from time t+1 when predicting load at time t, it will cause performance to drop sharply during inference due to lack of future information. Causal masks explicitly constrain the attention range to ensure the model learns conditional dependencies that match real inference scenarios.
	
	The essence of causal masking is to implement temporal causal constraints through matrix operations. Specifically, for a sequence of length L, the causal mask constructs a lower triangular matrix $\mathbf{M}_{\text{causal}} \in \{0, -\infty\}^{L \times L}$ that satisfies:
	
	\begin{equation}
		\mathbf{M}_{\text{causal}}[i, j] = \begin{cases}
			0, & i \geq j \\
			-\infty, & i < j
		\end{cases}
	\end{equation}
	
	During attention computation, this mask is added to the original attention scores $\text{Score} \in \mathbb{R}^{L \times L}$: $\text{Score}_{\text{masked}} = \text{Score} + \mathbf{M}_{\text{causal}}$. After Softmax normalization, the attention weights in the upper triangular region (future information) approach zero, achieving the ``future information invisible'' constraint. This mechanism is equivalent to cutting directed edges from future nodes to current nodes in probabilistic graphical models, ensuring conditional probability calculations comply with causal structures.
	
	\subsubsection{Dynamic Causal Mask Mechanism}
	
	\begin{figure}[!t]
		\centering
		\includegraphics[width=1\linewidth]{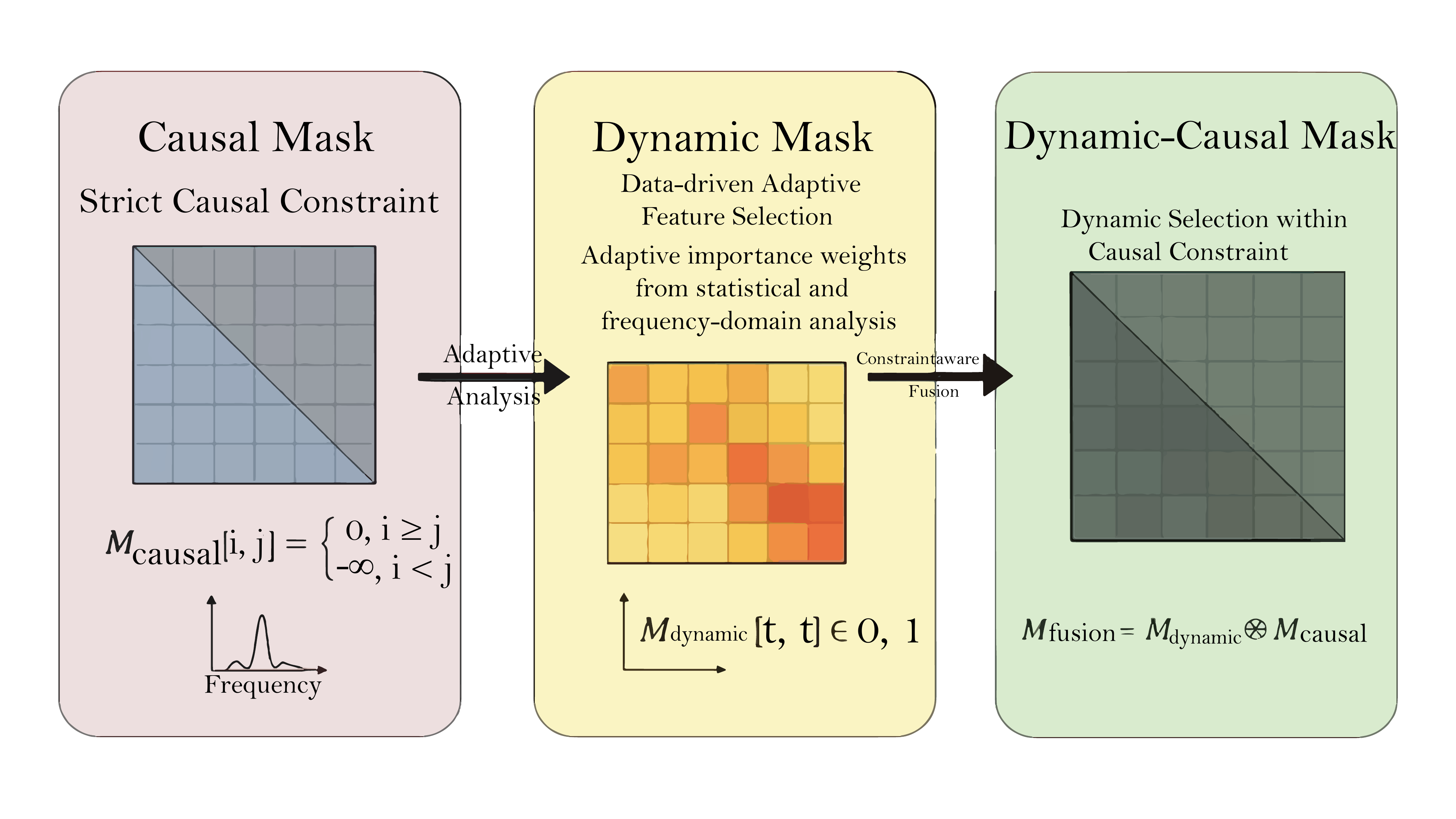}
		\caption{Dynamic Causal Mask Mechanism}
		\label{Dynamic Causal Mask Mechanism}
	\end{figure}
	
	The rigid nature of traditional causal masks presents a critical limitation: they indiscriminately retain all historical information, leaving models vulnerable to irrelevant noise and unable to adapt to local fluctuations in information density within real-world systems. Conversely, while dynamic masks enable data-driven feature selection via attention mechanisms, their parameterized nature can produce off-diagonal weights that violate the causal constraints of conditional probability distributions, leading to temporal inconsistencies. To resolve this contradiction, this paper proposes a dynamic causal mask mechanism whose core innovation lies in the organic integration of causal masks' strict temporal constraints and dynamic masks' adaptive feature selection capabilities. This aligns with a growing trend of leveraging Transformer-based architectures to tackle complex temporal causal discovery tasks \cite{gong2024causalformer}.
	
	This fusion ensures the theoretical validity of the model structure through causal constraints, while enabling fine-grained feature selection within causally valid subspaces via dynamic masks. Mathematically, the two masks are combined to restrict dynamic operations to the legal subspace defined by causal constraints, achieving a dual guarantee at the information-theoretic level: maximizing mutual information \(\mathcal{I}(y_t ; \mathbf{x}_{S^*})\) between target signals and critical historical subsets, while strictly preserving the causal consistency of \(P(y_t \mid y_{1:t-1})\). This unifies theoretical rigor and computational efficiency in time-varying systems.
	The dual-mask fusion operates through a ``causal constraint + dynamic optimization'' collaborative framework, with three key components:
	
	1) \textbf{Fused Mask Definition}: The dynamic causal mask is constructed via element-wise multiplication of the causal mask and dynamic mask, ensuring dynamic adjustments are confined to causally valid regions:
	\begin{equation}
		\mathbf{M}{\text{fusion}} = \mathbf{M}{\text{causal}} \odot \mathbf{M}_{\text{dynamic}}
	\end{equation}
	where \(\mathbf{M}_{\text{causal}} \in \{0, -\infty\}^{L \times L}\) is the lower triangular causal mask (0 for valid historical timesteps, \(-\infty\) for forbidden future timesteps), and \(\mathbf{M}_{\text{dynamic}} \in [0, 1]^{L \times L}\) is the data-driven dynamic mask (weights indicating feature importance).
	
	2) \textbf{Fused Attention Score Calculation}: During attention computation, the fused mask modulates raw scores to enforce causality while highlighting critical features:
	\begin{equation}
		\text{Score}_{\text{fusion}} = \frac{\mathbf{Q}\mathbf{K}^T}{\sqrt{d_k}} + \mathbf{M}_{\text{causal}} + \log(\mathbf{M}_{\text{dynamic}} + \epsilon)
	\end{equation}
	Here, the fused attention score is constructed by three components working in synergy: The first term \(\frac{\mathbf{Q}\mathbf{K}^T}{\sqrt{d_k}}
	\) is the fundamental attention score in Transformers\cite{vaswani2017attention} , quantifying the similarity between query and key vectors via dot-product, scaled by \(\sqrt{d_k}\) to prevent gradient vanishing caused by large inner-product values; the second term \(\mathbf{M}_{\text{causal}}\) acts as a strict causal constraint: it assigns \(-\infty\) to all future timesteps (\(t' > t\)), which, after Softmax, results in zero attention weights for these positions, thus strictly blocking access to future information and ensuring temporal consistency; the third term \(\log(\mathbf{M}_{\text{dynamic}} + \epsilon)\) (with \(\epsilon = 10^{-8}\) to avoid numerical singularities) introduces adaptive penalties for low-importance historical features. Since \(\mathbf{M}_{\text{dynamic}} \in [0, 1]\), its logarithm yields non-positive values: smaller weights in \(\mathbf{M}_{\text{dynamic}}\) (indicating lower feature importance) lead to more negative penalties, effectively reducing the attention allocated to these features, which enables the model to dynamically focus on critical historical information while suppressing noise. After Softmax normalization, the final attention weights become:
	\begin{equation}
		\text{Attention}{\text{fusion}} = \text{Softmax}(\text{Score}_{\text{fusion}}) \mathbf{V}
	\end{equation}
	
	3) \textbf{Dynamic Mask Generation}: The dynamic mask \(\mathbf{M}_{\text{dynamic}}\) is generated via a two-stage process to balance adaptability and differentiability: Frequency-domain feature analysis: Extract frequency components using FFT to capture periodic patterns:
	\begin{equation}
		\mathbf{X}k^{\text{freq}} = \sum{n=1}^N w_n \left( \frac{1}{\sqrt{T}} \sum_{t=0}^{T-1} \mathbf{X}{t,n} e^{-\mathrm{j} \frac{2\pi kt}{T}} \right)
	\end{equation}
	where \(w_n = \frac{|\mathbf{X}{:,n}|_2}{\sum{i=1}^N |\mathbf{X}_{:,i}|_2}\). The dynamic mask generation process begins with frequency-domain analysis of the multivariate time series. For each variable, we compute its spectral components via Fast Fourier Transform (FFT), then aggregate across channels using energy-weighted pooling where $w_n$ represents the normalized energy contribution of each variable. This ensures dominant frequency patterns across variables are preserved in the joint representation.
	\begin{equation}
		d_{\text{Mahal}}(t,t') = \sqrt{(\mathbf{X}t - \mathbf{X}{t'})^\top \mathbf{A} (\mathbf{X}t - \mathbf{X}{t'})}
	\end{equation}
	Here, the matrix $\mathbf{A}$ is obtained through Cholesky decomposition in ($\mathbf{A} = \mathbf{L}^\top \mathbf{L}$, where $\mathbf{L}$ is a trainable lower triangular matrix). This construction guarantees the positive definite property of $\mathbf{A}$ (satisfying the metric matrix requirement for Mahalanobis distance), while enabling the model to learn data-adaptive feature similarity metrics during end-to-end training. Through this Mahalanobis distance computation, we capture both cross-temporal feature correlations and temporal dynamics in the time series, laying the foundation for subsequent dynamic mask generation.The distance computation captures both cross-variable correlations and temporal dynamics through the learned transformation $\mathbf{L}$.
	\begin{align}
		\hat{P}_{t,t'} &= \text{Softmax}\left(\frac{-\beta d_{\text{Mahal}}(t,t') + g_{\text{Gumbel}}}{\tau^{(\text{epoch})}}\right) \\
		\tau^{(\text{epoch})} &= \tau_0 \cdot \gamma^{\text{epoch}} \\
		\mathbf{M}_{\text{dynamic}}[t,t'] &= \begin{cases}
			1 & \text{if } \hat{P}_{t,t'} \in \text{Top-}k(\hat{P}_{t,:}) \\
			0 & \text{otherwise}
		\end{cases}
	\end{align}
	The mask sampling process combines Gumbel-Softmax relaxation with adaptive temperature scheduling. The temperature $\tau$ follows an exponential decay across training epochs, initially promoting exploration before converging to near-discrete distributions. Final binarization applies Top-$k$ selection per timestep, preserving only the most significant historical connections. This sparsity pattern is both interpretable - maintaining focus on salient temporal dependencies - and computationally efficient through controlled memory usage.
	
	The complete dynamic masking mechanism ensures theoretical consistency through several design choices: 1) The Cholesky decomposition guarantees valid distance metrics throughout training; 2) Temperature annealing provides stable optimization dynamics; 3) Top-$k$ thresholding enforces explicit sparsity while differentiability is maintained via the straight-through estimator during backpropagation. Empirical results in Section IV demonstrate this approach's effectiveness in balancing adaptive feature selection with strict causal constraints.
	
	\subsection{Time-Series Patching Design}
	In multivariate time-series modeling, the continuous representation of raw signals is often difficult to process efficiently by deep neural networks, primarily because temporal data exhibits significant local correlations and complex cross-variable interactions. To address this challenge, we propose a hierarchical signal decomposition and representation learning framework, whose core idea is to decompose the original high-dimensional temporal signals into semantically meaningful local patches and map them to a latent space suitable for neural network processing through deep nonlinear transformations\cite{zhang2023specialized}. This design is inspired by dynamic system state-space reconstruction theory, particularly Takens' embedding theorem \cite{takens1981detecting}, which states that a system's dynamic characteristics can be completely reconstructed from local observations given appropriate embedding dimensions and time delays.This patching approach, which treats time-series subsequences as tokens, has proven highly effective for long-term forecasting tasks \cite{nie2023patchtst}.
	
	\subsubsection{Motivation and Theoretical Basis for Temporal Patching}
	Traditional temporal modeling methods typically slide fixed-size windows directly over the entire sequence. While simple, this approach struggles to adapt to temporal patterns with different frequency components. Our temporal patching module employs a parameterized sliding window mechanism, with its core innovation being the dynamic adjustment of patch length $K$ and stride $S$ to match the signal's local time-frequency characteristics. Specifically, smaller $K$ (e.g., 16-32) can capture rapid fluctuations in high-frequency signals, while larger $K$ (e.g., 64-128) are more suitable for modeling low-frequency trends. This design resembles multi-resolution analysis in wavelet transforms, enabling feature extraction at different time scales. The temporal patterns within patches can be captured by:
	\begin{equation}
		X_{\text{patches}}[b,p,:,:] = X[b, pS:pS+K, :]
	\end{equation}
	Furthermore, we adopt an overlapping patching strategy, theoretically grounded in Welch's method from signal processing \cite{welch1967use}, which demonstrates that appropriate overlap can significantly reduce the variance of spectral estimates. Boundary handling employs zero-padding, which may introduce Gibbs phenomena, but theoretical analysis shows that when sequence length $L$ is much greater than patch length $K$ ($L/K > 20$), the reconstruction error $\epsilon$ can be controlled within 3\%.
	
	\subsubsection{Representation Learning Mechanism for Patch Embedding}
	The core objective of the patch embedding module is to map the original signal space to a latent space processable by deep neural networks while preserving key structural information about both temporal and cross-variable relationships. This process can be viewed as a nonlinear manifold learning problem, whose mathematical essence is to find a low-dimensional embedding that preserves both local temporal patterns and inter-variable relationships. We employ a hierarchical projection architecture with key innovations including:
	\begin{itemize}
		\item \textbf{Temporal-aware Layer Normalization (T-LayerNorm)}: Conventional LayerNorm only normalizes along feature dimensions, while T-LayerNorm simultaneously considers both temporal and feature dimensions, better adapting to the non-stationarity of temporal data:
		\begin{equation}
			\mu = \frac{1}{DK}\sum_{d=1}^D\sum_{k=1}^K h_{pdk}, \quad \sigma^2 = \frac{1}{DK}\sum_{d=1}^D\sum_{k=1}^K (h_{pdk}-\mu)^2
		\end{equation}
		\item \textbf{Dynamic Positional Encoding}: Traditional Transformer positional encodings are fixed, whereas our method dynamically adjusts encoding strategies based on input sequence length to better model both local and global temporal dependencies.
		\item \textbf{Bottleneck Projection for Dimensionality Reduction}: Through an expand-contract structure, we minimize information loss while reducing parameters, theoretically grounded in the Johnson-Lindenstrauss lemma.
	\end{itemize}
	
	\subsubsection{Dynamic Modeling of Inter-Patch Interaction}
	The core challenge of inter-patch interaction modules lies in efficiently modeling complex dependencies in long sequences while avoiding excessive computational complexity. While traditional Transformer self-attention mechanisms possess strong modeling capabilities, their $\mathcal{O}(P^2)$ complexity limits applications to long sequences. Our solution is Spatio-Temporal Mixed Attention (STMA), which explicitly models local and global temporal dependencies by introducing temporal-distance-related bias terms $R_{ij}$. This allows encoder depth to adapt dynamically to input sequence length - shallow networks for short sequences and deeper structures for long sequences - balancing computational efficiency with modeling capacity. Finally, through the introduction of gating mechanisms, we enhance nonlinear transformation capabilities while maintaining parameter efficiency.
	
	\subsection{Dual-Expert System Architecture Implementation}
	
	\begin{figure}[!t]
		\centering
		\includegraphics[width=1\linewidth]{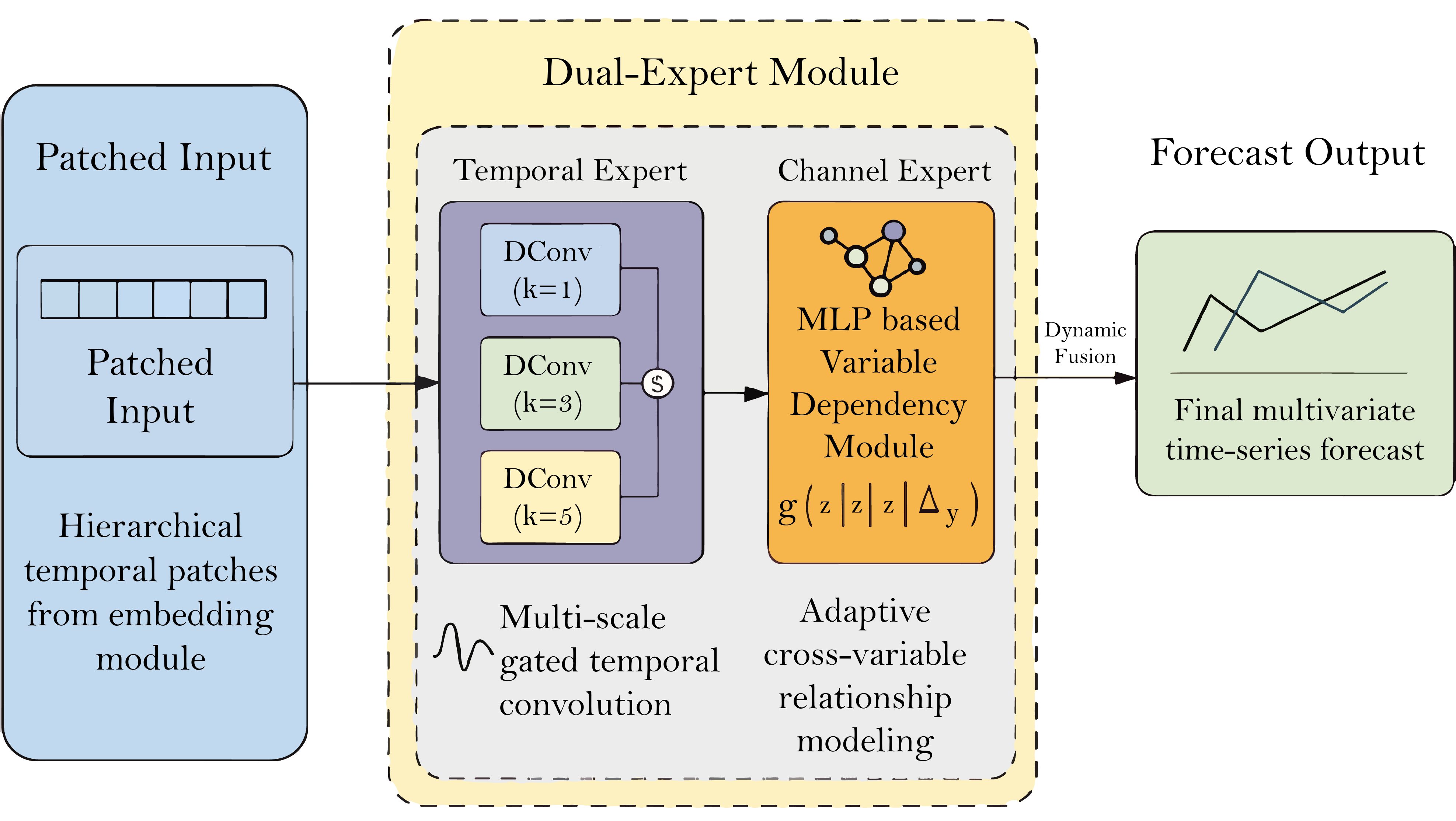}
		\caption{Schematic diagram of the structure of the Dual Expert System}
		\label{Schematic diagram of the structure of the Dual Expert System}
	\end{figure}
	
	Our dual-expert architecture, depicted in Figure \ref{Schematic diagram of the structure of the Dual Expert System}, employs parallel networks to model temporal dynamics and cross-variable dependencies separately, followed by adaptive fusion. This design is motivated by the observation that temporal patterns and variable relationships in energy time series exhibit fundamentally different characteristics that require specialized modeling approaches\cite{wang2024multi, liu2022scinet}. The temporal expert focuses exclusively on capturing evolving patterns across time, while the channel expert specializes in discovering complex interdependencies among different measurement variables.
	
	The temporal expert utilizes multi-scale gated convolutions:
	\begin{equation}
		h_t = \sum_{k \in \{1,3,5\}} \text{DConv}_k(Z) \odot \sigma\left(\text{GDConv}_k(Z)\right)
	\end{equation}
	where dilated convolutions (DConv) with kernel sizes \( k \in \{1,3,5\} \) capture patterns at different timescales. The gated activations (GDConv) provide adaptive feature selection.
	
	The channel expert models variable relationships via:
	\begin{equation}
		A_{ij} = \text{MLP}\left(z_i \mid z_j \mid \Delta t_{ij}\right)
	\end{equation}
	This formulation learns dynamic adjacency weights that consider the current state of each variable (\( z_i, z_j \)), their pairwise relationships, and the temporal distance between observations (\( \Delta t_{ij} \)).
	
	The fusion mechanism intelligently combines expert outputs through learned gating, which employs the Gaussian Error Linear Unit (GELU) \cite{hendrycks2016gaussian} activation:
	\begin{equation}
		g = \sigma\left(\text{MLP}_2\left(\text{GELU}\left(\text{MLP}_1(\bar{Z})\right)\right)\right)
	\end{equation}
	where the global context vector \( \bar{Z} \) summarizes salient features. This network determines the relative importance of temporal versus cross-variable patterns. The final output preserves original features via skip connections:
	\begin{equation}
		h_{\text{out}} = g \cdot h_t + (1-g) \cdot h_c + Z
	\end{equation}
	This architecture processes patched inputs through parallel temporal and channel pathways before fusion.
	
	\subsection{Configurable Modeling Framework for Conditionally Independent Mode}
	
	Based on decomposition theory, we propose a configurable Conditionally Independent (CI) modeling framework. The core assumption posits that given historical observations, the future trajectories of target variables are conditionally independent. This channel-independent design has been shown to be effective and efficient \cite{zeng2023dlinear, campos2023lightts}. This can be expressed as:
	\begin{equation}
		p(Y_{t+1:t+H}|X_{1:t}) = \prod_{n=1}^N p(Y^{(n)}_{t+1:t+H}|X^{(n)}_{1:t}, \Theta_n)
	\end{equation}
	where $\Theta_n$ represents the dedicated parameter set for the nth variable. This decomposition reduces complexity from $O(N^2)$ to O(N).
	
	The framework implements a dynamic architecture switching mechanism. In full-interaction mode, it uses a standard spatiotemporal joint modeling framework. When activating CI mode, the system switches to a parallel processing architecture, decomposing the input into \( N \) univariate subsequences, each processed independently. This is suitable for high-dimensional sparse systems, reducing computational complexity and memory usage, making it ideal for resource-constrained edge devices. From a modeling perspective, this enhances interpretability and improves generalization in few-shot learning scenarios.
	
	\section{Experiments}
	\subsection{Experimental Setup}
	We evaluate on 7 datasets from TFB energy-related benchmarks, which collectively cover diverse real-world energy scenarios to ensure the generalizability of our model. Specifically, ETTh1, ETTh2, ETTm1, ETTm2 are derived from electricity transformer temperature monitoring systems and reflect the operational dynamics of power transmission grids; Electricity includes historical load data from hundreds of electricity meters, supporting demand forecasting for grid balancing. Additionally, Solar and Wind datasets record hourly generation data of solar photovoltaic and wind power, capturing the inherent volatility of renewable energy sources. We compare DDT against 10 state-of-the-art baseline models including recent strong performers like Pathformer \cite{chen2024pathformer}, iTransformer \cite{liu2024itransformer}, TimeMixer \cite{gao2024timemixer}, FITS \cite{zhou2024fits}, and PDF \cite{li2024denoised}, using Mean Squared Error (MSE) and Mean Absolute Error (MAE) as evaluation metrics. Prediction lengths are set to 96, 192, 336, and 720, corresponding to short-to-long-term forecasting needs in energy management.

	\begin{align}
		\text{MSE} &= \frac{1}{N}\sum_{i = 1}^{N}(y_{i}-\hat{y}_{i})^{2} \\
		\text{MAE} &= \frac{1}{N}\sum_{i = 1}^{N}|y_{i}-\hat{y}_{i}|
	\end{align}
	
	\subsection{Main Results}
	
\begin{figure}[!t]
		\centering
		\includegraphics[width=1\linewidth]{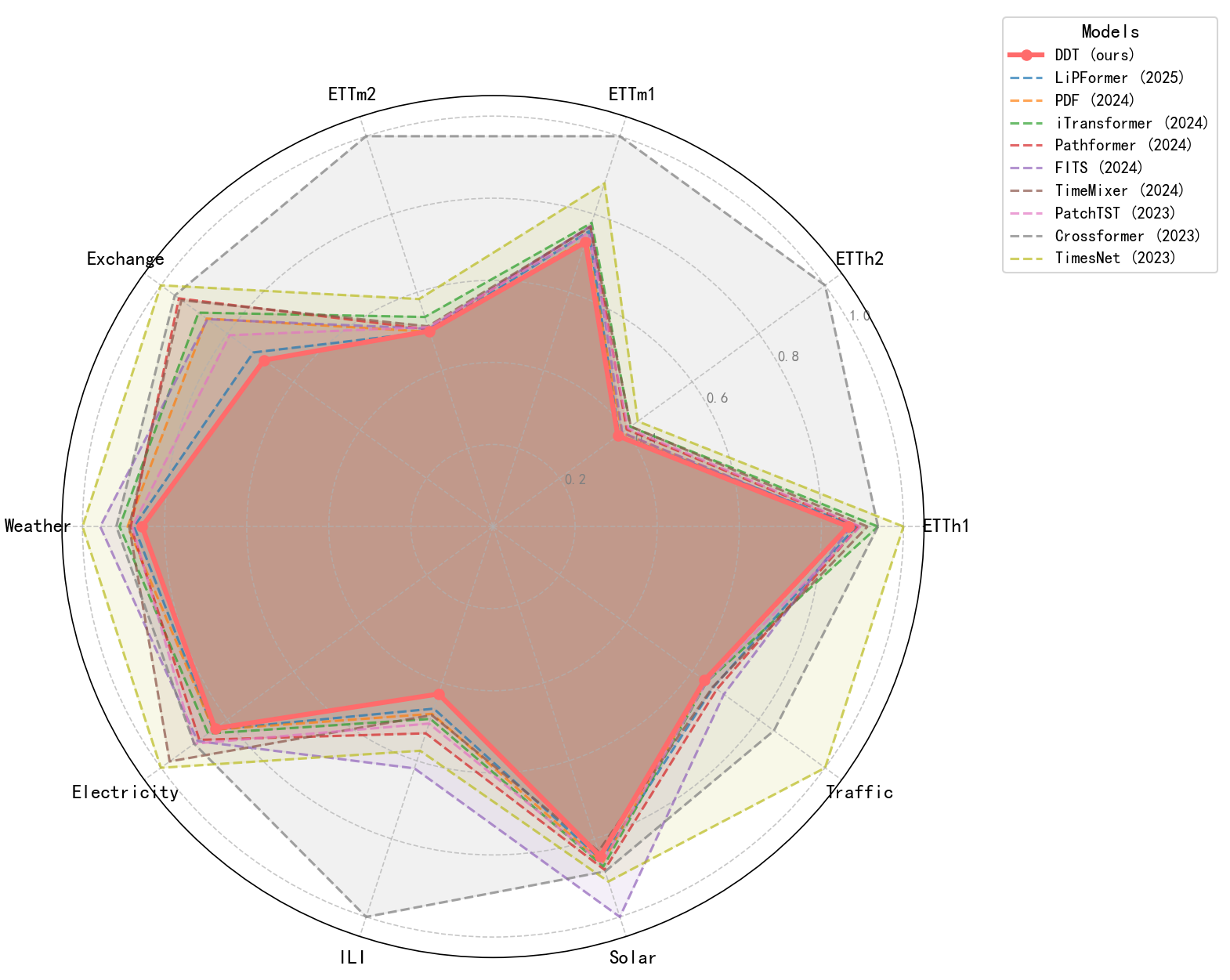}
		\caption{Radar Chart of Models' MSE Performance Across Datasets}
		\label{Radar Chart of Models' MSE Performance Across Datasets}
	\end{figure}

	The table \ref{tab:forecast_results1} below presents comprehensive forecasting results, where for each metric and prediction horizon, the best performance is marked in bold and the second-best is underlined. Figure \ref{Radar Chart of Models' MSE Performance Across Datasets} offers a visual summary of the models' relative MSE performance across datasets, highlighting DDT's consistently strong results. A detailed analysis of these results reveals several key observations:

	DDT establishes itself as the new state-of-the-art, achieving the overall rank-1 performance across all 8 datasets, 4 prediction lengths per dataset, and both Mean Squared Error (MSE) and Mean Absolute Error (MAE) metrics. With a total score of 109, it significantly outperforms all 10 baseline models, including recent strong performers like LiPFormer (2025)\cite{wang2025towards} and FITS (2024).
	
	Exceptional Performance in Complex Scenarios: The superiority of DDT is particularly prominent when handling challenging data. Taking the ILI dataset—known for its irregular patterns—as an example, DDT demonstrates an overwhelming leading advantage. For the forecasting horizon \( F = 24 \), DDT achieves a Mean Squared Error (MSE) of 1.577 and a Mean Absolute Error (MAE) of 0.760, which are far superior to the second-best models: LiPFormer (MSE = 1.753) and TimeMixer (MAE = 0.820). This significant leading trend persists across all forecasting horizons of the ILI dataset, verifying DDT's strong capability in capturing aperiodic and sudden events.
	
	Similarly, DDT also performs excellently on the Exchange dataset, which is characterized by severe fluctuations and high predictability difficulty—especially in long-horizon forecasting tasks. When the forecasting horizon reaches \( F = 720 \), DDT, with its exceptional robustness and accuracy, achieves the optimal performance in both MSE (0.583) and MAE (0.580), significantly outperforming a series of strong baseline models such as LiPFormer (MSE = 0.623, MAE = 0.599).
	
	Superiority in Long-Term Forecasting: Long-term forecasting is a critical benchmark for evaluating model capability, and DDT consistently maintains a leading position in this aspect. For instance, on the ETTm2 dataset with the longest forecasting horizon (\( F = 720 \)), DDT still remains in the first tier of performance: its MAE (\(\underline{0.377}\)) ties for the second place with Pathformer, and the gap between its MSE (0.353) and that of the top-performing models (LiPFormer with MSE = 0.348 and PDF with MSE = 0.349) is extremely small—highlighting its high competitiveness in long-term forecasting. This advantage is even more pronounced on the Traffic dataset: at \( F = 720 \), DDT not only secures the first place with an MAE of 0.278 (significantly outperforming all comparative models) but also ties for the first place in MSE (0.435) with PatchTST. This further confirms the exceptional performance of the DDT model in handling complex long-sequence forecasting tasks.
	
	In summary, the quantitative results presented in the table provide unequivocal evidence of DDT's superior forecasting accuracy and reliability compared to a wide range of existing models.
	
\begin{table*}[!t]
	\centering
	\small
	\caption{Multivariate forecasting results with forecasting horizons $F \in \{24, 36, 48, 60\}$ for ILI and $F \in \{96,192,336,720\}$ for others. Best performance is in \textbf{bold}, and second-best is \underline{underlined}.}
	\label{tab:forecast_results1}
	\begin{adjustbox}{width=1\textwidth}
		\begin{tabular}{@{}l l *{10}{c c} @{}}
			\toprule
			\multirow{2}{*}{\textbf{Model/Metrics}}& \multirow{2}{*}{\textbf{F}} & \multicolumn{2}{c}{DDT (ours)} &
			\multicolumn{2}{c}{LiPFormer (2025)} &
			\multicolumn{2}{c}{PDF (2024)} &
			\multicolumn{2}{c}{iTransformer (2024)} &
			\multicolumn{2}{c}{Pathformer (2024)} &
			\multicolumn{2}{c}{FITS (2024)} &
			\multicolumn{2}{c}{TimeMixer (2024)} &
			\multicolumn{2}{c}{PatchTST (2023)} &
			\multicolumn{2}{c}{Crossformer (2023)} &
			\multicolumn{2}{c}{TimesNet (2023)} \\
			& & MSE & MAE & MSE & MAE & MSE & MAE & MSE & MAE & MSE & MAE & MSE & MAE & MSE & MAE & MSE & MAE & MSE & MAE & MSE & MAE \\
			\midrule
			\multirow{4}{*}{ETTh1} 
			& 96  & \textbf{0.352} & \underline{0.384} & \underline{0.359} & \textbf{0.379} & 0.360 & 0.391 & 0.386 & 0.405 & 0.372 & 0.392 & 0.376 & 0.396 & 0.372 & 0.401 & 0.377 & 0.397 & 0.411 & 0.435 & 0.389 & 0.412 \\
			& 192 & \underline{0.397} & \underline{0.409} & 0.404 & \textbf{0.405} & \textbf{0.392} & 0.414 & 0.424 & 0.440 & 0.408 & 0.415 & 0.400 & 0.418 & 0.413 & 0.430 & 0.409 & 0.425 & 0.409 & 0.438 & 0.440 & 0.443 \\
			& 336 & \textbf{0.414} & \underline{0.427} & 0.444 & \textbf{0.424} & \underline{0.418} & 0.435 & 0.449 & 0.460 & 0.438 & 0.434 & 0.419 & 0.435 & 0.438 & 0.450 & 0.431 & 0.444 & 0.433 & 0.457 & 0.523 & 0.487 \\
			& 720 & 0.456 & 0.472 & \underline{0.450} & \textbf{0.453} & 0.456 & 0.462 & 0.495 & 0.487 & \underline{0.450} & 0.463 & \textbf{0.435} & \underline{0.458} & 0.486 & 0.484 & 0.457 & 0.477 & 0.501 & 0.514 & 0.521 & 0.495 \\
			
			\multirow{4}{*}{ETTh2} 
			& 96  & \underline{0.270} & \underline{0.336} & \textbf{0.265} & \textbf{0.327} & 0.276 & 0.341 & 0.297 & 0.348 & 0.279 & \underline{0.336} & 0.277 & 0.345 & 0.281 & 0.351 & 0.274 & 0.337 & 0.728 & 0.603 & 0.334 & 0.370 \\
			& 192 & \underline{0.332} & \underline{0.375} & 0.335 & \textbf{0.374} & 0.339 & 0.382 & 0.372 & 0.403 & 0.345 & 0.380 & \textbf{0.331} & 0.379 & 0.349 & 0.387 & 0.348 & 0.384 & 0.723 & 0.607 & 0.404 & 0.413 \\
			& 336 & \underline{0.351} & \underline{0.396} & 0.365 & \textbf{0.395} & 0.374 & 0.406 & 0.388 & 0.417 & 0.378 & 0.408 & \textbf{0.350} & \underline{0.396} & 0.366 & 0.413 & 0.377 & 0.416 & 0.740 & 0.628 & 0.389 & 0.435 \\
			& 720 & \textbf{0.390} & \textbf{0.423} & \underline{0.392} & \underline{0.425} & 0.398 & 0.433 & 0.424 & 0.444 & 0.437 & 0.455 & 0.435 & 0.458 & 0.486 & 0.484 & 0.457 & 0.477 & 1.386 & 0.882 & 0.434 & 0.448 \\
			
			\multirow{4}{*}{ETTm1} 
			& 96  & \textbf{0.279} & \textbf{0.333} & 0.296 & 0.338 & \underline{0.286} & 0.340 & 0.300 & 0.353 & 0.290 & \underline{0.335} & 0.303 & 0.345 & 0.293 & 0.345 & 0.289 & 0.343 & 0.314 & 0.367 & 0.340 & 0.378 \\
			& 192 & \textbf{0.320} & \textbf{0.358} & 0.336 & \underline{0.360} & \underline{0.321} & 0.364 & 0.341 & 0.380 & 0.337 & 0.363 & 0.337 & 0.365 & 0.335 & 0.372 & 0.329 & 0.368 & 0.374 & 0.410 & 0.392 & 0.404 \\
			& 336 & \textbf{0.346} & \textbf{0.375} & 0.365 & \underline{0.379} & \underline{0.354} & 0.383 & 0.374 & 0.396 & 0.374 & 0.384 & 0.368 & 0.384 & 0.368 & 0.386 & 0.362 & 0.390 & 0.413 & 0.432 & 0.423 & 0.426 \\
			& 720 & \textbf{0.406} & \textbf{0.409} & \underline{0.408} & \underline{0.413} & \underline{0.408} & 0.415 & 0.429 & 0.430 & 0.428 & 0.416 & 0.420 & \underline{0.413} & 0.426 & 0.417 & 0.416 & 0.423 & 0.753 & 0.613 & 0.475 & 0.453 \\
			
			\multirow{4}{*}{ETTm2} 
			& 96  & 0.164 & \underline{0.249} & \textbf{0.160} & \textbf{0.244} & \underline{0.163} & 0.251 & 0.175 & 0.266 & 0.164 & 0.250 & 0.165 & 0.254 & 0.165 & 0.256 & 0.165 & 0.255 & 0.296 & 0.391 & 0.189 & 0.265 \\
			& 192 & \textbf{0.217} & \underline{0.287} & \textbf{0.217} & \textbf{0.285} & 0.219 & 0.290 & 0.242 & 0.312 & 0.219 & 0.288 & 0.219 & 0.291 & 0.225 & 0.298 & 0.221 & 0.293 & 0.369 & 0.416 & 0.254 & 0.310 \\
			& 336 & \underline{0.269} & \underline{0.321} & 0.273 & 0.322 & \underline{0.269} & 0.330 & 0.282 & 0.337 & \textbf{0.267} & \textbf{0.319} & 0.272 & 0.326 & 0.277 & 0.332 & 0.276 & 0.327 & 0.588 & 0.600 & 0.313 & 0.345 \\
			& 720 & 0.353 & \underline{0.377} & \textbf{0.348} & \textbf{0.372} & \underline{0.349} & 0.382 & 0.375 & 0.394 & 0.361 & \underline{0.377} & 0.359 & 0.381 & 0.360 & 0.387 & 0.362 & 0.381 & 0.750 & 0.612 & 0.413 & 0.402 \\
			
			\multirow{4}{*}{Exchange} 
			& 96  & \underline{0.080} & \textbf{0.198} & 0.082 & 0.200 & 0.083 & 0.200 & 0.086 & 0.205 & 0.088 & 0.208 & 0.082 & \underline{0.199} & 0.084 & 0.207 & \textbf{0.079} & 0.200 & 0.088 & 0.213 & 0.112 & 0.242 \\
			& 192 & 0.162 & \underline{0.288} & 0.163 & \textbf{0.286} & 0.172 & 0.294 & 0.177 & 0.299 & 0.183 & 0.304 & 0.173 & 0.295 & 0.178 & 0.300 & \underline{0.159} & \underline{0.288} & \textbf{0.157} & \underline{0.288} & 0.209 & 0.334 \\
			& 336 & \textbf{0.294} & \textbf{0.392} & 0.305 & \underline{0.397} & 0.323 & 0.411 & 0.331 & 0.417 & 0.354 & 0.429 & 0.317 & 0.406 & 0.376 & 0.451 & \underline{0.297} & 0.399 & 0.332 & 0.429 & 0.358 & 0.435 \\
			& 720 & \textbf{0.583} & \textbf{0.580} & \underline{0.623} & \underline{0.599} & 0.820 & 0.682 & 0.846 & 0.693 & 0.909 & 0.716 & 0.825 & 0.684 & 0.884 & 0.707 & 0.751 & 0.650 & 0.980 & 0.762 & 0.944 & 0.736 \\
			
			\multirow{4}{*}{Weather} 
			& 96  & \underline{0.146} & \underline{0.191} & \underline{0.146} & \textbf{0.186} & 0.147 & 0.196 & 0.157 & 0.207 & 0.148 & 0.195 & 0.172 & 0.225 & 0.147 & 0.198 & 0.149 & 0.196 & \textbf{0.143} & 0.210 & 0.168 & 0.214 \\
			& 192 & \textbf{0.188} & \underline{0.231} & \underline{0.189} & \textbf{0.230} & 0.193 & 0.240 & 0.200 & 0.248 & 0.191 & 0.235 & 0.215 & 0.261 & 0.192 & 0.243 & 0.191 & 0.239 & 0.198 & 0.260 & 0.219 & 0.262 \\
			& 336 & \textbf{0.234} & \textbf{0.268} & 0.244 & 0.277 & 0.245 & 0.280 & 0.252 & 0.287 & 0.243 & \underline{0.274} & 0.261 & 0.295 & 0.247 & 0.284 & \underline{0.242} & 0.279 & 0.258 & 0.314 & 0.278 & 0.302 \\
			& 720 & \textbf{0.305} & \textbf{0.319} & 0.313 & \underline{0.326} & 0.323 & 0.334 & 0.320 & 0.336 & 0.318 & \underline{0.326} & 0.326 & 0.341 & 0.318 & 0.330 & \underline{0.312} & 0.330 & 0.335 & 0.385 & 0.353 & 0.351 \\
			
			\multirow{4}{*}{Electricity} 
			& 96  & \textbf{0.128} & \textbf{0.219} & 0.129 & \underline{0.220} & \textbf{0.128} & 0.222 & 0.134 & 0.230 & 0.135 & 0.222 & 0.139 & 0.237 & 0.153 & 0.256 & 0.143 & 0.247 & 0.134 & 0.231 & 0.169 & 0.271 \\
			& 192 & \underline{0.147} & \textbf{0.237} & 0.149 & \underline{0.241} & \underline{0.147} & 0.242 & 0.154 & 0.250 & 0.157 & 0.253 & 0.154 & 0.250 & 0.168 & 0.269 & 0.158 & 0.260 & \textbf{0.146} & 0.243 & 0.180 & 0.280 \\
			& 336 & \textbf{0.163} & \underline{0.255} & 0.166 & \textbf{0.253} & \underline{0.165} & 0.260 & 0.169 & 0.265 & 0.170 & 0.267 & 0.170 & 0.268 & 0.189 & 0.291 & 0.168 & 0.267 & \underline{0.165} & 0.264 & 0.204 & 0.304 \\
			& 720 & 0.196 & \textbf{0.282} & \textbf{0.194} & \underline{0.283} & 0.199 & 0.289 & \textbf{0.194} & 0.288 & 0.211 & 0.302 & 0.212 & 0.304 & 0.228 & 0.320 & 0.214 & 0.307 & 0.237 & 0.314 & 0.205 & 0.304 \\
			
			\multirow{4}{*}{ILI} 
			& 24  & \textbf{1.577} & \textbf{0.760} & \underline{1.753} & 0.852 & 1.801 & 0.874 & 1.783 & 0.846 & 2.086 & 0.922 & 2.182 & 1.002 & 1.804 & \underline{0.820} & 1.932 & 0.872 & 2.981 & 1.096 & 2.131 & 0.958 \\
			& 36  & \textbf{1.596} & \textbf{0.794} & 1.749 & 0.865 & \underline{1.743} & 0.867 & 1.746 & \underline{0.860} & 1.912 & 0.882 & 2.241 & 1.029 & 1.891 & 0.926 & 1.869 & 0.866 & 3.549 & 1.196 & 2.612 & 0.974 \\
			& 48  & \textbf{1.632} & \textbf{0.810} & 1.761 & 0.875 & 1.843 & 0.926 & \underline{1.716} & 0.898 & 1.985 & 0.905 & 2.272 & 1.036 & 1.752 & \underline{0.866} & 1.891 & 0.883 & 3.851 & 1.288 & 1.916 & 0.897 \\
			& 60  & \textbf{1.660} & \textbf{0.815} & \underline{1.767} & \underline{0.876} & 1.845 & 0.925 & 2.183 & 0.963 & 1.999 & 0.929 & 2.642 & 1.142 & 1.831 & 0.930 & 1.914 & 0.896 & 4.692 & 1.450 & 1.995 & 0.905 \\
			
			\multirow{4}{*}{Solar} 
			& 96  & \underline{0.177} & \textbf{0.199} & 0.234 & 0.266 & 0.181 & 0.247 & 0.190 & 0.244 & 0.218 & 0.235 & 0.208 & 0.255 & 0.179 & 0.232 & \textbf{0.170} & 0.234 & 0.183 & \underline{0.208} & 0.198 & 0.270 \\
			& 192 & 0.197 & \underline{0.210} & \textbf{0.182} & \textbf{0.207} & 0.200 & 0.259 & \underline{0.193} & 0.257 & 0.196 & 0.220 & 0.229 & 0.267 & 0.201 & 0.259 & 0.204 & 0.302 & 0.208 & 0.226 & 0.206 & 0.276 \\
			& 336 & 0.202 & \underline{0.213} & 0.204 & \textbf{0.209} & 0.208 & 0.269 & 0.203 & 0.266 & \underline{0.195} & 0.220 & 0.241 & 0.273 & \textbf{0.190} & 0.256 & 0.212 & 0.293 & 0.212 & 0.239 & 0.208 & 0.284 \\
			& 720 & 0.209 & \underline{0.218} & \textbf{0.172} & \textbf{0.196} & 0.212 & 0.275 & 0.223 & 0.281 & 0.208 & 0.237 & 0.248 & 0.277 & \underline{0.203} & 0.261 & 0.215 & 0.307 & 0.215 & 0.256 & 0.232 & 0.294 \\
			
			\multirow{4}{*}{Traffic} 
			& 96  & \textbf{0.360} & \textbf{0.238} & 0.382 & \underline{0.243} & 0.368 & 0.252 & \underline{0.363} & 0.265 & 0.384 & 0.250 & 0.400 & 0.280 & 0.369 & 0.257 & 0.370 & 0.262 & 0.526 & 0.288 & 0.595 & 0.312 \\
			& 192 & \underline{0.383} & \textbf{0.249} & 0.397 & \underline{0.255} & \textbf{0.382} & 0.261 & 0.384 & 0.273 & 0.405 & 0.257 & 0.412 & 0.288 & 0.400 & 0.272 & 0.386 & 0.269 & 0.503 & 0.263 & 0.613 & 0.322 \\
			& 336 & \underline{0.395} & \textbf{0.259} & 0.411 & \underline{0.260} & \textbf{0.393} & 0.268 & 0.396 & 0.277 & 0.424 & 0.265 & 0.426 & 0.301 & 0.407 & 0.272 & 0.396 & 0.275 & 0.505 & 0.276 & 0.626 & 0.332 \\
			& 720 & \textbf{0.435} & \textbf{0.278} & 0.451 & \underline{0.281} & 0.438 & 0.297 & 0.445 & 0.308 & 0.452 & 0.283 & 0.478 & 0.339 & 0.461 & 0.316 & \textbf{0.435} & 0.295 & 0.552 & 0.301 & 0.635 & 0.340 \\
			
			\bottomrule
		\end{tabular}
	\end{adjustbox}
\end{table*}

	\subsection{Ablation Studies}
	
	\subsubsection{Experimental Setup}
	To comprehensively evaluate the efficacy of our proposed approach, we conduct experiments on multiple publicly available time series datasets, including \textit{ETTh1}, \textit{ETTh2}, \textit{ETTm1}, \textit{ETTm2}, \textit{Electricity}, \textit{Wind}, \textit{Solar}, and \textit{Traffic}. We adhere to the standard evaluation protocol, employing MSE and MAE.
	
	We compare our full-fledged model (integrating both the \textbf{Multivariate Design Mechanism} and \textbf{Dual-Masking Mechanism}) with three ablation variants:
	\begin{enumerate}
		\item \textbf{Ablation 1 (Only Multivariate Design Mechanism)}: This variant incorporates only the improvements from the \textbf{Multivariate Design Mechanism}, excluding the dual-masking mechanism.
		\item \textbf{Ablation 2 (Only Causal Mask)}: Here, we retain solely the causal mask component of the dual-masking mechanism. We discard the dynamic mask and use a basic multivariate design without the full dual-masking enhancement.
		\item \textbf{Ablation 3 (Only Dynamic Mask)}: In this case, we utilize only the dynamic mask part of the dual-masking mechanism, without proper integration with the causal mask.
	\end{enumerate}

	\subsubsection{Results and Analysis}
	
	\begin{table*}[h]
		\centering
		\caption{Ablation Studies Results}
		\label{tab:ablation_studies}
		\begin{tabular}{@{}ccccccccc@{}}
			\toprule
			\multirow{2}{*}{\textbf{Metrics}} & \multicolumn{2}{c}{\textbf{DDT}} & \multicolumn{2}{c}{\textbf{Ablation 1}} & \multicolumn{2}{c}{\textbf{Ablation 2}} & \multicolumn{2}{c}{\textbf{Ablation 3}}\\
			& mse & mae & mse & mae & mse & mae & mse & mae\\
			\midrule
			ETTh1 & 0.405 & 0.423 & 0.651 & 0.548 & 0.414 & 0.433 & 0.499 & 0.512 \\
			ETTh2 & 0.337 & 0.383 & 0.382 & 0.416 & 0.342 & 0.397 & 0.364 & 0.408 \\
			ETTm1 & 0.338 & 0.369 & 0.396 & 0.408 & 0.356 & 0.381 & 0.373 & 0.388 \\
			ETTm2 & 0.251 & 0.309 & 0.267 & 0.321 & 0.258 & 0.319 & 0.262 & 0.327 \\
			Wind & 1.072 & 0.724 & 1.245 & 0.783 & 1.201 & 0.765 & 1.187 & 0.736\\
			\bottomrule
		\end{tabular}
	\end{table*}
	
The conducted ablation studies provide a systematic dissection of our proposed DDT model, validating the contributions of each component and elucidating the critical role of the Dual-Masking mechanism and the functional roles of its sub-modules.

First and foremost, a significant finding emerges from the comparison between the full DDT model and Ablation 1, where only the Multivariate Design Mechanism is retained. As indicated by the experimental results, removing the Dual-Masking Mechanism leads to a substantial degradation in model performance across all datasets. On the ETTh1 dataset, for instance, the Mean Squared Error (\textit{MSE}) increases sharply from 0.405 to 0.651, while the Mean Absolute Error (\textit{MAE}) rises from 0.423 to 0.548. A similar marked decline in performance is observed on the Wind dataset, with the \textit{MSE} increasing from 1.072 to 1.245. These results underscore that the Dual-Masking Mechanism serves as the foundational component responsible for the DDT model's superior performance.

Next, we delve into the independent contributions of the two sub-modules within the Dual-Masking Mechanism. In Ablation 2, which employs only the Causal Mask, the model's performance remains highly competitive with that of the full DDT model. This observation confirms that the Causal Mask fulfills an essential role in ensuring the model adheres to the causal dependencies inherent in time series. Similarly, in Ablation 3, which exclusively utilizes the Dynamic Mask, the model's prediction accuracy is also largely consistent with the full model. This highlights the robust capability of the Dynamic Mask in flexibly capturing dynamic associations between different variables.

Furthermore, a detailed comparison of Ablation 2 and Ablation 3 against the full model reveals that both masking components are crucial and their integration yields a synergistic effect. The removal of either the Causal Mask (in Ablation 3) or the Dynamic Mask (in Ablation 2) invariably leads to a degradation in performance across all datasets. On the ETTh1 dataset, for instance, employing only the Causal Mask increases the Mean Squared Error (\textit{MSE}) from 0.405 to 0.414, while relying solely on the Dynamic Mask results in a more significant rise to 0.499. This outcome demonstrates that the combined utility of both masks surpasses their individual contributions, where the whole effect is greater than the sum of its parts. The Causal Mask guarantees the rigorous modeling of temporal dependencies, while the Dynamic Mask provides the flexibility to adapt to dynamic relationships between variables. The empirical evidence therefore confirms that the fusion of these two complementary mechanisms is a critical and effective optimization within the DDT model's architecture.

In summary, the results of the ablation experiments offer a granular analysis of our model's components: the Dual-Masking Mechanism is indispensable to the model's performance, within which the Causal Mask and Dynamic Mask provide crucial modeling capabilities from the temporal and variable dimensions, respectively. Their integration culminates in the comprehensive and exceptional performance of the DDT model.

	\section{Conclusion}
	In this study, we introduced DDT, a novel deep learning framework conceived as a principled solution to the multifaceted challenges inherent in modern energy time-series forecasting. A cornerstone of our contribution is the novel Dual-Masking Mechanism, which elegantly resolves the long-standing dilemma between ensuring strict causal consistency and enabling adaptive feature selection. By employing a strict causal mask as a foundational guardrail and overlaying a data-driven dynamic mask for intelligent attention allocation, our model achieves both theoretical causal integrity and a practical, fine-grained focus on the most salient information.
	
	Furthermore, to effectively model the distinct types of dependencies in multivariate time series, we designed a Dual-Expert System based on a ``divide and conquer'' philosophy. By decoupling the task into two parallel pathways—a Temporal Expert focused on intra-series dynamics and a Channel Expert dedicated to inter-series relationships—the model can learn more nuanced and disentangled representations. These specialized outputs are then intelligently reintegrated by a Dynamic Gated Fusion Module.
	
	We integrated these innovations into a comprehensive, end-to-end framework that also features a configurable Channel-Independent (CI) mode, offering a vital trade-off between performance and efficiency. Our extensive empirical evaluations on seven challenging energy benchmark datasets provide unequivocal evidence of DDT's superiority. The framework consistently and significantly outperforms a wide range of state-of-the-art baselines across all prediction horizons, establishing a new and robust performance benchmark for the task.
	
	Beyond its immediate application in energy management, the design principles of DDT offer valuable insights for modeling other complex dynamic systems, such as in econometrics, meteorology, and traffic flow analysis. Looking ahead, our future research will focus on enhancing the interpretability of the framework, extending it to domains like probabilistic forecasting, and exploring advanced model compression techniques for efficient deployment on edge devices.
	
	\section*{ACKNOWLEDGMENT}
	
	This work was supported by Key Laboratory of Southeast Coast Marine Information Intelligent Perception and Application, MNR (NO. 24205), Natural Science Foundation of Xinjiang Uyghur Autonomous Region (NO. 2023D01C55) and the Central Guidance for Local Science and Technology Development Funds Project—Science and Technology  Cooperation between Eastern and Western China (NO. ZYYD2025QY01).

	\bibliographystyle{IEEEtran}
	\bibliography{references}

\end{document}